# The Self-Driving Car: Crossroads at the Bleeding Edge of Artificial Intelligence and Law

Scott McLachlan[1,2,4], Evangelia Kyrimi[2], Kudakwashe Dube[3,4], Norman Fenton[2], and Burkhard Schafer[1]

[1]Script Centre, Edinburgh Law School, University of Edinburgh, UK
[2]Risk and Information Management, Electronic Engineering and Computer Science, Queen Mary University of London, UK
[3]School of Fundamental Sciences, Massey University, NZ
[4]Health Informatics and Knowledge Engineering research (HiKER) Group

**Abstract**
Artificial intelligence (AI) features are increasingly being embedded in cars and are central to the operation of self-driving cars (SDC). There is little or no effort expended towards understanding and assessing the broad legal and regulatory impact of the decisions made by AI in cars. A comprehensive literature review was conducted to determine the perceived *barriers, benefits and facilitating factors* of SDC in order to help us understand the suitability and limitations of existing and proposed law and regulation. (1) existing and proposed laws are largely based on claimed benefits of SDV that are still mostly speculative and untested; (2) while publicly presented as issues of assigning blame and identifying *who pays* where the SDC is involved in an accident, the barriers broadly intersect with almost every area of society, laws and regulations; and (3) new law and regulation are most frequently identified as the primary factor for enabling SDC. Research on assessing the impact of AI in SDC needs to be broadened beyond negligence and liability to encompass barriers, benefits and facilitating factors identified in this paper. Results of this paper are significant in that they point to the need for deeper comprehension of the broad impact of all existing law and regulations on the introduction of SDC technology, with a focus on identifying only those areas truly requiring ongoing legislative attention.

*The major difference between a thing that might go wrong and a thing that cannot possibly go wrong is that when a thing that*

*cannot possibly go wrong goes wrong it usually turns out to be impossible to get at and repair.*

Douglas Adams

*Guess we have some improvements to make before production haha*[1].

Elon Musk
(Tesla CEO)

Kenji Urada was a maintenance worker at Kawasaki's factory in Akashi, Japan. He unintentionally hit the 'on switch' as he mounted the safety barrier to visually check a large malfunctioning automotive gear processing robot and became pinned between the robot's arm and the machine itself as the arm swung around. This incident was further complicated by a lack of training within the plant that meant other workers were unable to deactivate the robot and rescue him. Urada had the unfortunate distinction of having been publicly proclaimed in the mainstream media as the first human killed by a robot[2]. However, it was later found that UPI's 1981 report had been incorrect and Urada was not the first man killed by a robot. That unenviable title belonged to Robert Williams, a Ford Motor Company factory worker killed by the one-tonne autonomous transfer cart of a five-storey parts retrieval robot at Ford's Detroit plant on January 25th, 1979[3]. It is ironic that both gentlemen who have held the title *first-man-killed-by-a-robot* died in factories belonging to companies that, among other things, produced motor vehicles.

# 1. Introduction

Most vehicles manufactured in the last three decades contain software in a myriad of small dedicated computers known as *electronic control units* (ECU)[4]. ECU underpin the operation of many individual vehicle subsystems [1]. While software in non-self-driving cars (non-SDC) The Society of Automotive Engineers (SAE) describe automation on an escalating scale of levels[5] from those that assist and enhance safe operation of the vehicle controlled by the human driver[6] at Level 0, through to *self-driving cars* (SDC) that incorporate intelligent software capable of fully subsuming the driver's role in navigation and control in all road, weather and traffic conditions at Level 5 [2, 3]. Many current vehicles are capable of tasks at Levels 1 and 2, which includes *automated driver assistance systems* (ADAS) that enable *automated lane centring* and *adaptive cruise control*. A very small number of vehicles are also capable of Level 3 functionality using features that enable autonomous control in limited circumstances like slow-moving traffic jams[7]. However, laws and road rules in some jurisdictions still limit the use of Level 3 functionality. Given that multiple companies are currently developing and testing SDC technology at higher levels[8], the fully automated self-driving cars of Knight Rider (KITT), Total Recall (The Johnny Cab) and the voice-activated *Auto Mode* seen in Demolition Man, Timecop and i-Robot are now less *futuristic science fiction* and more *near-term reality*.

[1] https://twitter.com/elonmusk/status/1198090787520598016

[2] https://www.upi.com/Archives/1981/12/08/Robot-kills-man/2127376635600/

[3] $10Million awarded to Family of U.S. Plant Worker Killed by Robot. Ottowa Citizen. August 11, 1983. p.14. https://news.google.com/newspapers?id=7KMyAAAAIBAJ&pg=3301,87702

[4] These include: *engine control unit* (ECU) that controls the air-fuel mix, timing and fuel injection systems; *powertrain control module* that provides a control interface between the engine and transmission systems; *body control module* (BCM) that monitors and controls the electronic accessories in the vehicle such as power windows, central door locks and air conditioning; the *vehicle entertainment system* (VES) that operates the stereo head unit, maps and GPS, CD player and in some cases DVD video player; and even the digital *instrument clusters* of most modern cars.

[5] https://www.sae.org/news/press-room/2018/12/sae-international-releases-updated-visual-chart-for-its-%E2%80%9Clevels-of-driving-automation%E2%80%9D-standard-for-self-driving-vehicles

[6] For example, the blind spot detection, lane departure warning and automated braking solutions included in most vehicles manufactured during the last 1-2 decades.

[7] Tesla's *Autopilot*, BMW's *Driver Assistant Professional* and Mercedes Benz's *traffic-jam assist*.

[8] Including: Tesla's *Autopilot*, BMW's *Driver Assistant Professional*, Google Alphabet's *Waymo*, General Motor's *Cruise*, Lyft's *Self-driving lab*, Uber's *Aurora*, and Mercedes Benz's *Intelligent Drive*.

The regulatory environment for motor vehicles is complex: governing how motor vehicles are designed, manufactured, advertised for sale, product standards, maintenance requirements, ownership, insurance and driver licensure[9]. However, aside from conferences[10], public debates[11] and much literature focused largely on the issue of apportioning liability when SDC are involved in traffic accidents, the impact of the entire accumulation of existing law on SDC, and alternately of SDC on the application and operation of those laws, appears to be less extensively considered and we contend may not be fully understood. This work reviews and presents an overview of recent literature in the legal domain that discusses autonomous vehicle adoption and regulation, and is primarily intended to stimulate further discussion and broader inquiry in this area.

# 2. Background

The goal to replicate human actions using automated control systems, even in motor vehicles[12], is not new [4]. The first autopilot for planes was developed only nine years after the Wright Brothers inaugural flight at Kitty Hawk and could perform take-off, landing and maintaining cruising altitude [5, 6]. Two years after first conceiving of his autopilot invention, Lawrence Sperry successfully demonstrated this gyroscope-based autopilot device to the French military at a 1914 aviation safety contest in Paris [5], and since then the term *autopilot* has become synonymous for any technology that automates human control functions[13]. Modern food and medicine production is now conducted in predominantly automated factories [7, 8], and many recent developments in the fields of AI and *machine learning* (ML) predominantly seek approaches for automating tasks naturally performed by humans [9].

While seemingly unrelated, the development paths of aeroplanes and motor vehicles have shared a strong connection. Technologies developed for airplanes have found application on motor vehicles many times. Anti-lock braking systems (ABS) in motor vehicles arose in the 1920's as an invention to prevent aircraft wheels from locking up during rapid deceleration on low-traction surfaces like icy runways; and after Bosch received a patent for the first ABS controller in 1936, ABS were regularly built into large passenger planes [10, 11]. While other computer-controlled ABS systems had been installed in vehicles since 1971, it was in 1978 that Mercedes released the S-class sedan that was the first production motor vehicle with the Bosch-designed ABS brake package [12]. Collision avoidance technology from which the aircraft industry's now ubiquitous *Traffic Collision Avoidance Systems* (TCAS) arose was first proposed in the late 1950's; and like aircraft versions, vehicular collision avoidance systems demonstrated in Japan in 1991 incorporated vision systems, radar, inertial heading sensors, inter-vehicle communications and more recently *global positioning system* (GPS) technologies [14, 15]. It is collision avoidance technology, and especially GPS, that have had the most profound enabling effect on SDC. Figure 1 shows a timeline of the history for autopilot and SDC technologies, indicating where GPS entered into both development streams.

Consequently, it is not so much a matter of if, but rather when SDC will become ubiquitous on our highways.

[9] Figure 10 provided later in this paper provides a sample of the various law and regulations that govern motor vehicles in two jurisdictions: Australia and the United Kingdom.

[10] Keynote addresses and breakout sessions at the EMEA Claims Conference 2018 focused on Autonomous cars and liability: https://www.swissre.com/dam/jcr:cdb7c0c4-9919-4177-8299-4e0ea16c9e03/3+-+AUTONOMOUS+CARS+AND+LIABILITY+break+out+session+presentation.pdf

[11] Northwestern University Law School ran a public debate on the 18th November, 2020 called *Autonomous Systems Failures: Who is Legally and Morally Responsible?* https://www.law.northwestern.edu/student-life/events/eventsdisplay.cfm?i=176039&c=2

[12] One of the first demonstrations of the concept of self-driving vehicles was seen in the General Motors 1939 New York World's Fair exhibit *Futurama*, wherein the concept of automated vehicles that could maintain their lane on the highway and be self-guided using radio signals from a traffic controller was presented to the general public.

[13] The Merriam-Webster dictionary defines autopilot as both a noun and a verb: *a device for automatically steering ships, aircraft and spacecraft* AND as *the automatic control provided by such a device*.

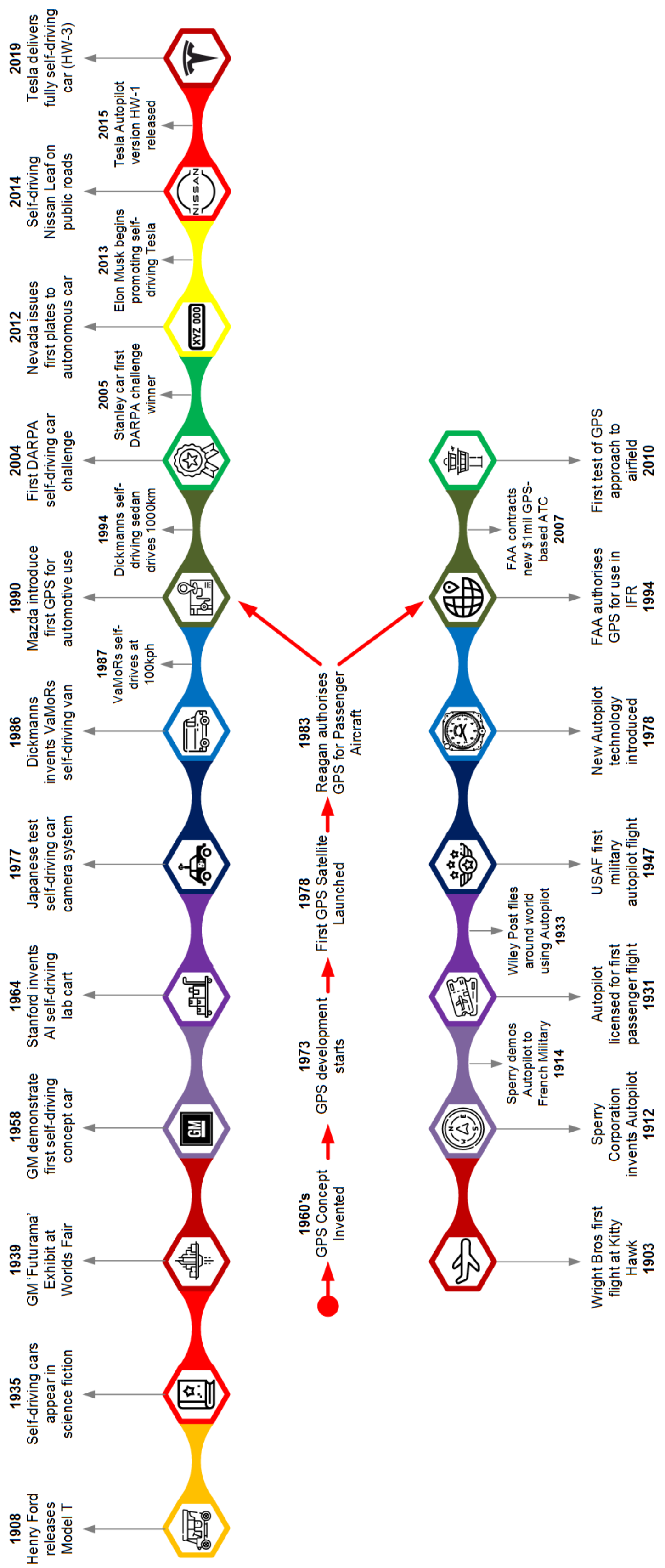

*Figure 1: Combined History of SDC and Autopilot Technologies*

***The History of SDC:*** One of the earliest examples of driverless cars was described in a science fiction story published in the May 1935 edition of the magazine *Wonder Stories*. Called "The Living Machine", that story was written by David H. Keller and incorporated the concept of SDC as central characters; describing the significant changes SDC brought to the lives of future humans[14] [16]. At the 1939 New York World's Fair the *General Motors* (GM) 'Futurama' exhibit proposed centrally radio-controlled self-driving cars would dominate the modern highway [17]. However, it was not until 1958 that GM were able to demonstrate a working prototype capable of autonomous driving on only a 400-foot stretch of Nebraska Highway equipped with special wiring embedded in the road surface [17]. Then, in 1977 the Japanese Tsubuka Mechanical Engineering Laboratory demonstrated a fully self-driving vehicle capable of driving at up to 20mph using two cameras to follow white markers painted on the road [18]. The next significant step forward for SDC came via Ernst Dickmanns, a German scientist who proposed the minimum five functions for an autonomous land vehicle[15], and demonstrated a self-driving delivery van called VaMoRs that was autonomously capable of negotiating unpopulated highways at up to 100kph [19]. In 1994 Dickmanns and his team went on to install an upgraded version of the VaMoRs 4-D machine vision system into twin Mercedes Benz S-class sedans that then drove autonomously for over 1000kms on motorways around Paris, France [20]. In 2012 the Nevada Department of Motor Vehicles issued the first registration to an autonomous vehicle from Google that used LiDAR [21], which was closely followed by Elon Musk's announcement that Tesla would eschew expensive LiDAR for cheaper camera-based sensors to provide autonomous driving features to his entire range of Tesla electric vehicles [22]. In 2019 Tesla delivered Autopilot version 3 (HW-3), and while their website describes this technology as *fully self-driving*, technology journalists say that it is not fully autonomous and still sits firmly at level 2 on SAE's Levels of Driver Automation[16] [23]. To date, Tesla's Autopilot represents the closest technology to true SDC that is commercially available to the general public. However, competing manufacturer's systems like BMW's *Driver Assistant Professional* (DAP) have potential to match or better Autopilot in the coming year.

***Technology enabling the SDC:*** The most significant advances in SDC-enabling technology have come in the form of advanced driver-assist systems (ADAS). Some authors see ADAS as electronic tools to assist drivers, while others focus almost exclusively on ADAS that are safety solutions capable of reducing traffic accidents and fatalities [24-27]. More correctly, ADAS represent groups of individual systems that each improve or automate some existing function of the driving process performed by the human driver. The standard most often relied on for classifying and describing ADAS is the Society for Automotive Engineers (SAE) Automated Driving Levels (SAE-ADL) [28, 29]. The six-level SAE-ADL taxonomy describe ADAS systems between the human driver and the automated system based on allocation of the driving subtask that the ADAS function operates on and the degree to which that subtask is automated by the ADAS: essentially, who is responsible for instant control of the vehicle. Levels 0-3 are differentiated by who controls two factors: (1) vehicle motion; and (2) the object event detection and response (OEDR) activities - which the SAE-ADL refer to collectively as the *dynamic driving task* (DDT). Levels 4 & 5 are differentiated based on whether the automation feature is capable of performing its subtask/s unrestricted by conditions imposed on or arising out of *geographic*, *road*, *environment*, *traffic*, *speed* or other limitations - which the SAE-ADL refer to as the *operational design domain* (ODD). Level 4 allows for automation constrained by one or more of these limitations at which point the driver must, or has the option to, take control of the vehicle, while Level 5 allows for complete system autonomy. Table 1 provides a list of existing common ADAS, describes their function and operation and classification based on SAE-ADL.

---

[14] "Old people began to cross the continent in their own cars. Young people found the driverless car admirable for petting. The blind for the first time were safe. Parents found they could more safely send their children to school in the new car than in the old cars with a chauffeur." - Extract from *The Living Machine* by David H. Keller.

[15] Dickmanns' minimum five functions required for autonomous mobility are: (1) detection of the road and its parameters like curvature, lane width, surface state and so on; (2) recognition of the vehicle's own relative position on the road; (3) correct interpretation of traffic information like road markings, signs and local warnings; (4) safe and efficient control of the vehicle within the parameters of 1-3); and (5) an easy-to-use interface for switching between autonomous and human driving modes.

[16] A Level 2 vehicle provides automated steering, braking and acceleration along with ADAS that in only limited circumstances and locations and under supervision of a human driver includes lane centering and adaptive cruise control. SAE's six levels of Driver Automation can be found here: https://www.sae.org/news/press-room/2018/12/sae-international-releases-updated-visual-chart-for-its-%E2%80%9Clevels-of-driving-automation%E2%80%9D-standard-for-self-driving-vehicles

*Table 1: ADAS Technologies*

| ADAS | | DESCRIPTION | SAE Level |
|---|---|---|---|
| ABS | Anti-lock Braking System | Avoids wheel lock-up and loss of control during full-force braking. Uses wheel speed sensors in a feedback system to regulate brake pressure on individual wheels as they begin to skid. | 1 |
| ACC | Adaptive Cruise Control | Maintains desired speed while also ensuring a safe distance between the vehicle and the vehicle ahead. Uses mid- and long-range radar and forward-looking cameras to detect vehicles ahead and a feedback system to control braking and engine acceleration. | 1 |
| ANS | Automotive Navigation System | Provides location and turn-by-turn directions. Uses digital mapping tools and global positioning (GPS) signals. | 0 |
| AP | Automated Parking | Capable of identifying suitable parking spaces and taking control of some or all of the functions to set the gear selector, steer, accelerate and brake in order to park the vehicle. Uses radar-based parking sensors and cameras and control mechanisms for braking, steering and acceleration (where supported). | 1-2 |
| BSD | Blind Spot Detection | Provides visual and/or audible warnings when another vehicle is located in or crosses into the blind spot of your vehicle. Uses short-range radar sensors. | 0 |
| BUC | Back Up Camera | Automatically engages when the driver selects the reverse gear. Uses rear-facing camera. | 0 |
| DDD | Driver Drowsiness Detection | Provides audible and sometimes tactile alerts in situations where an algorithm believes the driver may be drowsy or otherwise inattentive. Uses motion sensors, internal driver-facing cameras and data on recent tactile inputs such as steering or indicator use. | 0 |
| ESC | Electronic Stability Control | Provides assistance to maintain vehicle stability to avoid oversteer and understeer through selective braking on a per-wheel basis. Uses accelerometers and gyroscopic sensors, wheel speed sensors, driver input sensors and information regarding engine rpm and torque. | 1 |
| FCM | Forward Collision Mitigation | Also known as Automated Emergency Braking System (AEBS), FCM is an upgrade on FCW that identifies situations where the driver has not reacted and autonomously applies the brakes. Uses the mid-range radar and forward-facing cameras and control mechanisms for braking. | 1 |
| FCW | Forward Collision Warning | Uses audible and visual warnings to alert the driver to an imminent frontal collision. Uses mid-range radar and forward-facing cameras. | 0 |
| HUD | Heads Up Display | Reduces the need for looking away from the road, displays essential information such as current speed, posted speed limit, current gear and turn-by-turn directions on windshield in front of driver. Uses input from multiple systems including speed sensors, TSR and ANS. | 0 |
| LDW | Lane Departure Warning | Provides audible and sometimes tactile warning if the driver accidentally wanders across the boundary of the current lane. Uses the forward-facing camera to detect the position of lane markings. | 0 |
| LKC | Lane Keeping and Centring | An extension of LDW that can prevent unintentional lane departure and return the vehicle to the centre of the current lane. Uses the forward-facing camera and a feedback system to apply suitable control inputs to the steering wheel. | 1 |
| NV | Night Vision | Uses fixed light infra-red (FLIR) thermal imaging cameras to enable vision of obstacles in very low light conditions. | 0 |
| PS | Parking Sensors | Provide audible alerts to assist in avoiding obstacles at very low speeds. Uses ultrasonic sensors positioned on the front and rear bumpers. | 0 |
| TCS | Traction Control System | Prevents the wheels from slipping by cutting torque and keeping the vehicle stable. Uses ESC sensor system. | 1 |
| TSR | Traffic Sign Recognition | Identifies and alerts the driver to the current speed and other posted road rules. Uses forward-facing cameras attached to the windshield and often integrate with information from ANS. | 0 |

# 3. Method

The objective of this work is to carry out an exhaustive literature search of recent papers that satisfied the following characteristics:

1. SDC was the main topic;
2. Discussion of SDC implementation with at least some discourse on:
    a. Barriers impeding adoption of SDC,
    b. Benefits that may be realised when SDC can be ubiquitous, or
    c. Facilitating factors that may increase the probability of SDC adoption.
3. Discussion of approaches and/or limitations of current legal frameworks for SDC;

4. Proposal/s for new legislation or policies, or adaptation of existing law, that would be beneficial to implementation of SDC.

## 3.1 Search Terms

A search of the PubMed, HeinOnline, ScienceDirect, Scopus, DOAJ, and Elsevier databases was conducted. Preliminary eligibility for inclusion existed where the title, subject keywords or abstract uses at least one option from each of the three keyword groups identified in the following general search query:

```
"[(Self-driving) OR (Autonomous) OR (Automated)]
AND
[(Vehicle) OR (Car)]
AND
[(Law) OR (Legislation)]"
```

The initial literature search returned 1395 papers. After excluding papers published prior to 2016, those not published in English, or those whose content did not directly address legal or policy issues, there were 153 papers remaining.

# 4. Results

## 4.1 Literature search and collection results

The 153 papers that remained after the initial screening were weighted using the points in Table 2:

*Table 2: Literature Characteristics Weighting Chart*

| Characteristic | Points |
|---|---|
| SDC was the main topic | 1 |
| Law was a main or predominate topic | 1 |
| Included 1 benefit, OR | 1 |
| Included 2 or more benefit | 2 |
| Included 1 barrier, OR | 1 |
| Included 2 or more barriers | 2 |
| Included 1 facilitating factor, OR | 1 |
| Included 2 or more facilitating factors | 2 |
| Discussed 1 current law or regulation, OR | 1 |
| Discussed 2 or more current laws or regulations | 2 |
| Discussed the need for 1 required/future law or regulation, OR | 1 |
| Discussed the need for 2 or more required/future laws or regulations | 2 |

If a paper were to receive the highest score in each section it was possible to achieve a total weighting of 12. Papers had to receive a weighting of at least 8 to demonstrate sufficient relevance and remain in the review. Each of the remaining 48 papers was given a full read which resulted in an additional 7 being removed: four where law or legislation had only been mentioned in the abstract or introduction but was not analysed or discussed within the manuscript, one where vehicles, AI and autonomous driving were mentioned in the introduction but where the potentially arbitrary application of algorithms and intelligence in decision-making was the focus rather than SDC [37], and two which focused on quantitative psychological, philosophical and ethical evaluations of SDC decisions in accidents using the *trolley problem*[17] [38, 39]. As shown in Figure 2, this resulted in a final collection of 41 papers for use in this review.

[17] The trolley problem is a series of thought experiments involving hypothetical ethical dilemmas in which the subject person is told to imagine they stand at a track junction and must choose whether to allow a runaway trolley to remain on the main line and kill five people, or use a lever to divert it onto side track and kill one person. The dilemma is escalated wherein the subject is told additional facts, for example: that the single victim is a parent, pregnant mother or singularly skilled life-saving surgeon while the five victims are dock workers or homeless.

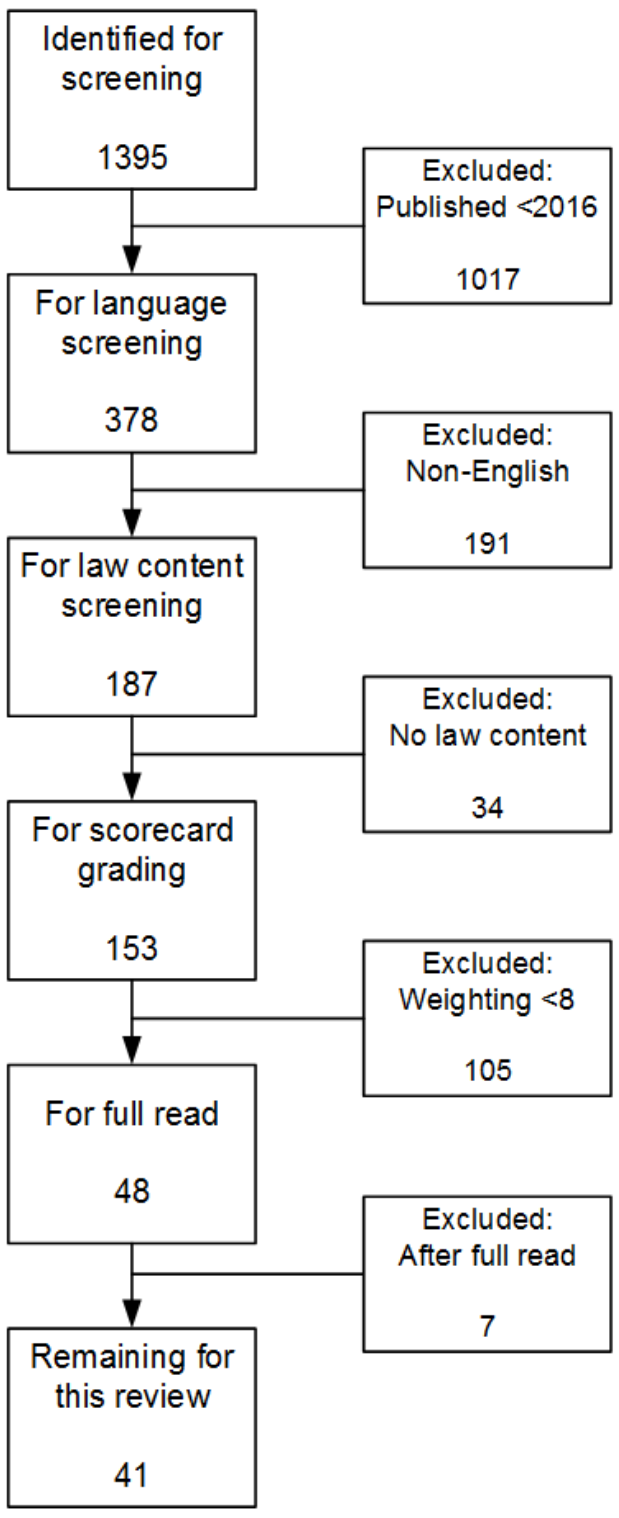

*Figure 2: Literature Search Outcome*

## 4.2 Nomenclature for SDC

A range of terms were used to represent SDC in the literature. Use of each term was highlighted, recorded and the context evaluated. Figure 3 shows that three nouns were identified. *Vehicle* and *car* were regularly used, with 24 authors using both interchangeably - at times even within the same paragraph. One work used the term *automobile* alongside both of the more common terms [40].

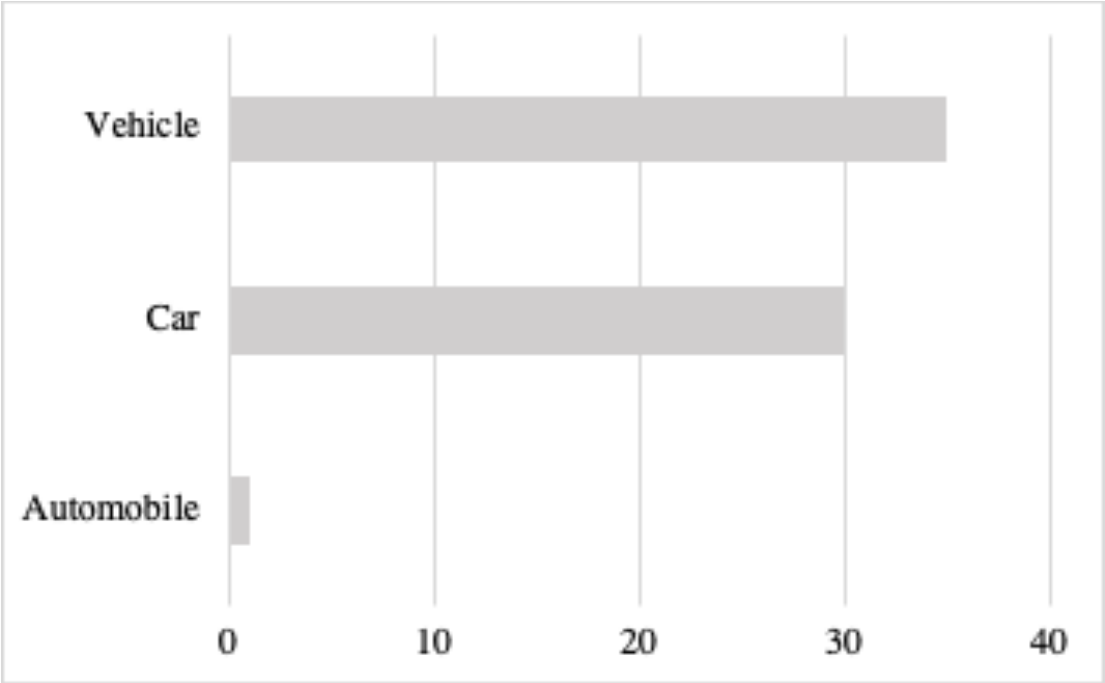

*Figure 3: Nouns used to describe SDC*

Five adjectives routinely described the artificial intelligence component of the SDC. Figure 4 shows *autonomous* and *self-driving* are common to more than three quarters of the works. The terms *inter/connected*, *driverless* and *automated* were observed where authors discussed respectively: (a) internetworking of SDC often with high-speed data networks like 5G to share location, speed, driving and lane-change intentions and other traffic-related data; (b) laws adverse to the implementation of SDC that require human drivers or *hands on the wheel*; or (c) vehicle functions not completely autonomous but which automated some function of driving such as steering, braking, or driving control in slow moving congested traffic.

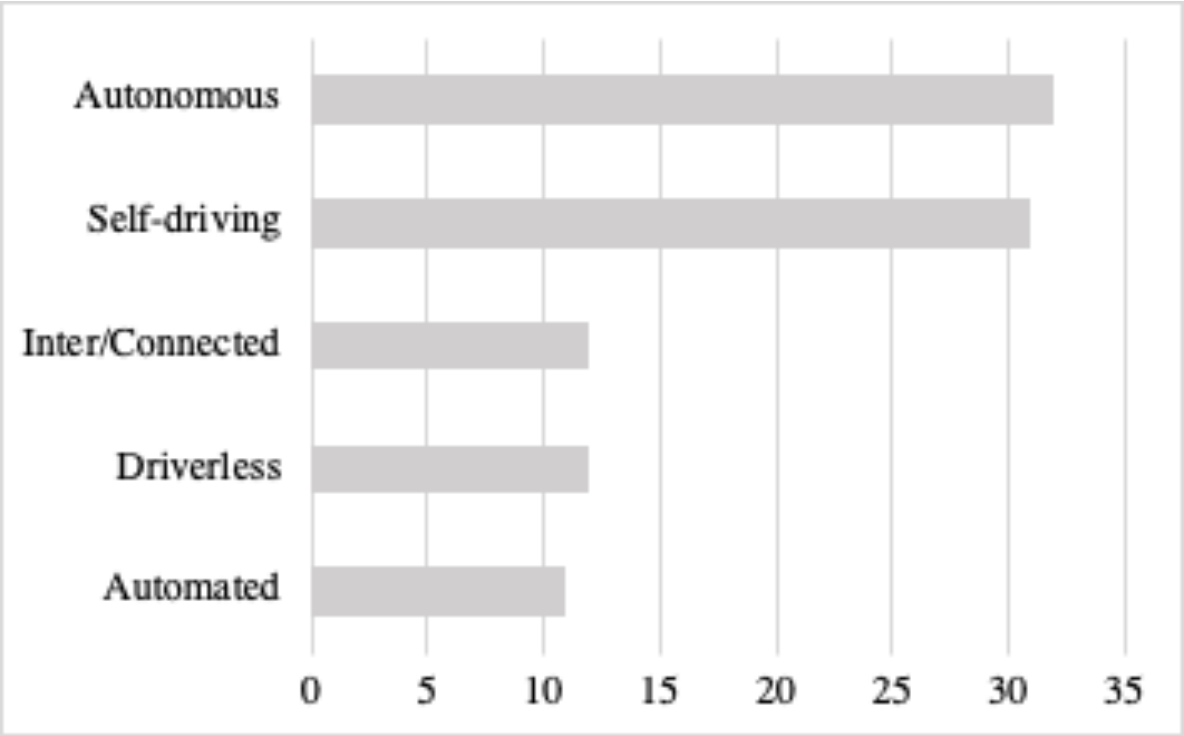

*Figure 4: Adjectives describing the Artificial Intelligence component of SDC*

The terms authors constructed using these nouns and adjectives and their incidence in the literature is shown in Table 3. Two constructed terms were observed more frequently: *Autonomous vehicles* and *self-driving* cars. However, few papers (*n* = 7) consistently used a single term to describe SDC, such that it was common for up to five different terms to be used in a paper. In two cases the constructed term for SDC even changed between the title to the first words of the paper [41, 40]. Several papers identified and discussed the lack of appropriate or consistent definitions in legislation for important concepts for vehicles generally but SDC in particular, such as for the terms *automated, autonomous*, and *driver* [18, 42, 43]. Definitions in legislation determine legal meaning and help ensure legal certainty for terms that can sometimes have ambiguous or competing *common use* meanings [44]. The ambiguity in legislative meanings for terms could also be seen in the clear inconsistency in meanings ascribed to the adjectives used to construct terms in the literature, such that in many cases words with different technical meanings like *automated* and *autonomous* were being used synonymously.

*Table 3: Constructed terms for SDC*

| | Connected and Autonomous Vehicles (CAV) | Connected and Automated Vehicle (CAV) | Automated, Autonomous and Connected Vehicles | Autonomous Intelligent Cars | Autonomous Vehicles | Semi-autonomous Vehicles | Semi-autonomous Cars | Automated Vehicles | Autonomous Cars | Autonomous Self-driving Vehicles | Driverless Cars | Driverless Vehicles | Autonomous or Self-driving Cars | Automated or Self-driving Vehicles | Automated Self-driving Cars | Self-driving Cars | Self-driving Vehicles | Connected Cars | **TOTAL TERMS USED** |
|---|---|---|---|---|---|---|---|---|---|---|---|---|---|---|---|---|---|---|---|
| **TOTAL** | **4** | **2** | **1** | **1** | **27** | **2** | **1** | **6** | **3** | **2** | **7** | **2** | **3** | **1** | **1** | **23** | **7** | **1** | |
| Gaeta (2019) [45] | | | | | ✓ | | | | | | ✓ | | | | | ✓ | | | **3** |
| Roth (2016) [46] | | | | | ✓ | | | | | | | | | | | | | | **1** |
| Kennedy (2019) [47] | ✓ | | | | | | | | | | | | | | | | ✓ | | **2** |
| Pearl (2018) [48] | | | | | ✓ | | ✓ | ✓ | ✓ | | ✓ | | | | | | | | **5** |
| Barrett (2017) [49] | | | | | ✓ | | | ✓ | | | | | | | | ✓ | | ✓ | **4** |
| Schellekens (2018) [50] | | | | | | | | | | | | | | | | | ✓ | | **1** |
| Ljungholm (2019) [51] | | | | | ✓ | | | | | | | | | | | | | | **1** |
| Rojas-Rueda et al (2020) [52] | | | | | ✓ | | | | | | | | | | | ✓ | | | **2** |
| Dixon et al (2020) [53] | | | | | | | | | | | | | | | ✓ | ✓ | | | **2** |
| Lohmann (2016) [40] | | | | | | | | | | | | | | ✓ | | ✓ | ✓ | | **3** |
| Taeihagh et al (2019) [54] | | | | | ✓ | | | | | | | ✓ | | | | | | | **2** |
| Browne (2017) [55] | | | | | | | | | | | | | ✓ | | | ✓ | | | **2** |
| Channon et al (2019) [56] | ✓ | | | | | | | | | | | | | | | | | | **1** |
| Brodsky (2016) [18] | | | | | ✓ | | | | ✓ | | ✓ | | | | | ✓ | | | **4** |
| Mardirossian (2020) [57] | | | | | ✓ | | | | ✓ | | | | | | | | | | **2** |
| Punev (2020) [58] | | | | | ✓ | | | | | | | | | | | | | | **1** |
| Roe (2019) [59] | | | | | ✓ | | | | | | ✓ | | | | | | | | **2** |
| Cohen et al (2018) [60] | | | | | ✓ | | | | | | | | | | | ✓ | | | **2** |
| Pittman et al (2019) [43] | | | | | | | | | | | | | | | | ✓ | | | **1** |
| Eastman (2016) [61] | | | | | ✓ | | | | | | | | | | | | ✓ | | **2** |
| Sun (2020) [62] | | | | | ✓ | | | | | | | | | | | ✓ | | | **2** |
| Leiman (2020) [63] | | | | | ✓ | | | ✓ | | | | | | | | | | | **2** |
| Rzeczycka et al (2019) [64] | | | | | ✓ | | | | | ✓ | | | | | | ✓ | | | **3** |
| Livak (2019) [41] | | | | | | | | | | | | | | | | ✓ | ✓ | | **2** |
| Mircica (2019) [65] | ✓ | | | | | | | | | | | | | | | ✓ | | | **2** |
| Cheng (2019) [42] | | | | | ✓ | | | | | ✓ | | | | | | ✓ | | | **3** |
| Mladenovic et al (2020) [66] | | ✓ | | | | | | | | | | | | | | | | | **1** |
| Marynowski (2019) [67] | | | | | ✓ | | | | | | ✓ | | | | | ✓ | | | **3** |
| Kouroutakis (2020) [68] | | | | | ✓ | | | | | | ✓ (red) | | | | | ✓ (red) | | | **3** |
| JonesDay (2017) [69] | | | ✓ | | ✓ | | | ✓ | | | | | | | | | | | **3** |
| Ryan (2020) [70] | | | | | | | | | | | | | | | | ✓ | ✓ | | **2** |
| Herbert (2017) [71] | | | | | ✓ | | | ✓ | | | | | | | | | | | **2** |
| Hamilton et al (2019) [72] | ✓ | | | | | | | ✓ | | | | | | | | | | | **2** |
| Coca-Vila (2018) [73] | | | | | | | | | | | | | ✓ (red) | | | ✓ (red) | | | **2** |
| Stilgoe (2018) [74] | | | | | ✓ (red) | | | | | | ✓ (red) | | ✓ | | | ✓ (red) | | | **4** |
| Greenbatt (2016) [75] | | | | | ✓ | | | | | | | | | | | ✓ | | | **2** |
| De Bruyne et al (2018) [76] | | ✓ | | | ✓ | | | | | | | | | | | ✓ | | | **3** |
| Zomarev et al (2020) [77] | | | | | ✓ | | | | | | | | | | | ✓ | ✓ | | **3** |
| Bruning et al (2017) [78] | | | | ✓ | ✓ | | | | | | | | | | | | | | **2** |
| Pearah (2017) [79] | | | | | | ✓ | | | | | | ✓ | | | | ✓ | | | **3** |
| Pollanen et al (2020) [80] | | | | | ✓ (red) | ✓ (red) | | | | | | | | | | | | | **2** |

**Legend:** ✓ = Author/s used these terms ✓ = Terms used alternately in same sentence or paragraph

## 4.3 The 3-D Taxonomy for SDC

ADAS support three primary functional domains: (i) *detection* of objects and potential hazards, (ii) *defence* of the vehicle, occupants and others through driver warnings and exercising control over the application of brakes; and (iii) augmenting *direction* by identifying the vehicle's current position and offering turn-by-turn navigation, or exercising control over steering to avoid hazards or self-park.

While some ADAS appear to only operate within the realm of a single functional domain, most occupy two domains and can be described on a spectrum based on the degree to which they operate within each domain. This paper presents this spectrum in the form of three domains as shown in Figure 1, which we allegorically call the *Gravity Well Diagram (GWD) of the 3-D Taxonomy for SDC*. We named Figure 5 the GWD because in Astrophysics, gravity wells, more commonly known as black holes, are the great attractors that draw matter to converge at the central event horizon which provides us with the allegory presented here. In the GWD the point of origin for each axis, where each domain is at its minimum, occurs on the outside. In our GWD the ambition to deliver operational SDC serves to motivate ADAS technology development along the axes toward this central goal and full automation of the SDC occurs at the apex of all three.

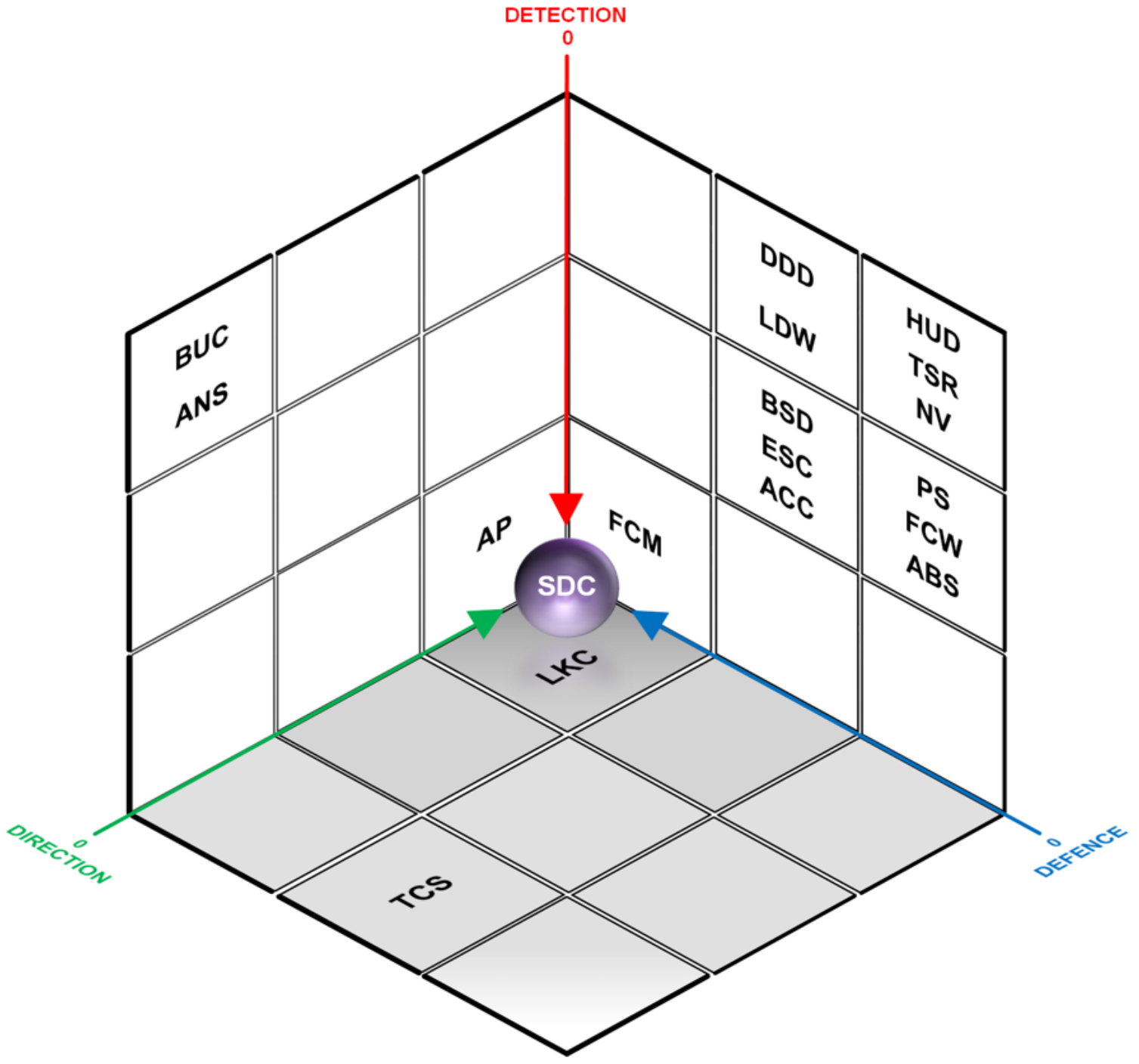

| | |
|---|---|
| **ABS** | Anti-lock Braking System |
| **ACC** | Adaptive Cruise Control |
| **ANS** | Automotive Navigation System |
| **AP** | Automated Parking |
| **BSD** | Blind Spot Detection |
| **BUC** | Back Up Camera |
| **DDD** | Driver Drowsiness Detection |
| **ESC** | Electronic Stability Control |
| **FCW** | Forward Collision Warning |
| **FCM** | Forward Collision Mitigation |
| **HUD** | Heads Up Display |
| **LDW** | Lane Departure Warning |
| **LKC** | Lane Keeping & Centring |
| **NV** | Night Vision |
| **PS** | Parking Sensors |
| **TCS** | Traction Control System |
| **TSR** | Traffic Sign Recognition |

*Figure 5: Gravity Well Diagram of the 3-D Taxonomy for SDC*

## 4.4 Benefits, barriers and facilitating factors

Benefits, barriers and facilitating factors were themes, each a collection of concepts, identified in every work. *Benefits* present as the positive outcomes authors propose that will result from ubiquitous adoption of the target item or intervention; for SDC these included reduced traffic congestion and increased productivity. Benefits are usually realised only after fixing one or more *barriers,* which present as obstacles preventing or limiting adoption, such as issues with assignment of liability and limitations inherent in existing legislation. Finally, *facilitating factors* are those activities that when successfully undertaken are able to overcome one or more existing barriers, and in doing so, thus promote successful adoption. An example might be enacting new legislation that enables real-world road testing of SDC.

### 4.4.1 Benefits

The potential for ubiquitous SDC adoption to benefit individuals, family groups and the wider community was a common theme. Listed in order of frequency in Figure 6, thirteen benefits were identified from the literature. That SDC would improve road safety was a strong claim in 63% of works, with authors believing *improved safety* would manifest in a reduction of traffic accidents (49%) and that fewer accidents would translate to fewer deaths (29%). Potential environmental benefits were discussed in 44% of works, where authors variously described *reduced individual vehicle ownership* [55, 79] and the interconnected nature of SDC to enable *ridesharing* [70, 77], *decrease congestion* [49, 46], *improve route planning* [40] and enable the ability to *safely draft off of each other* [55, 73].

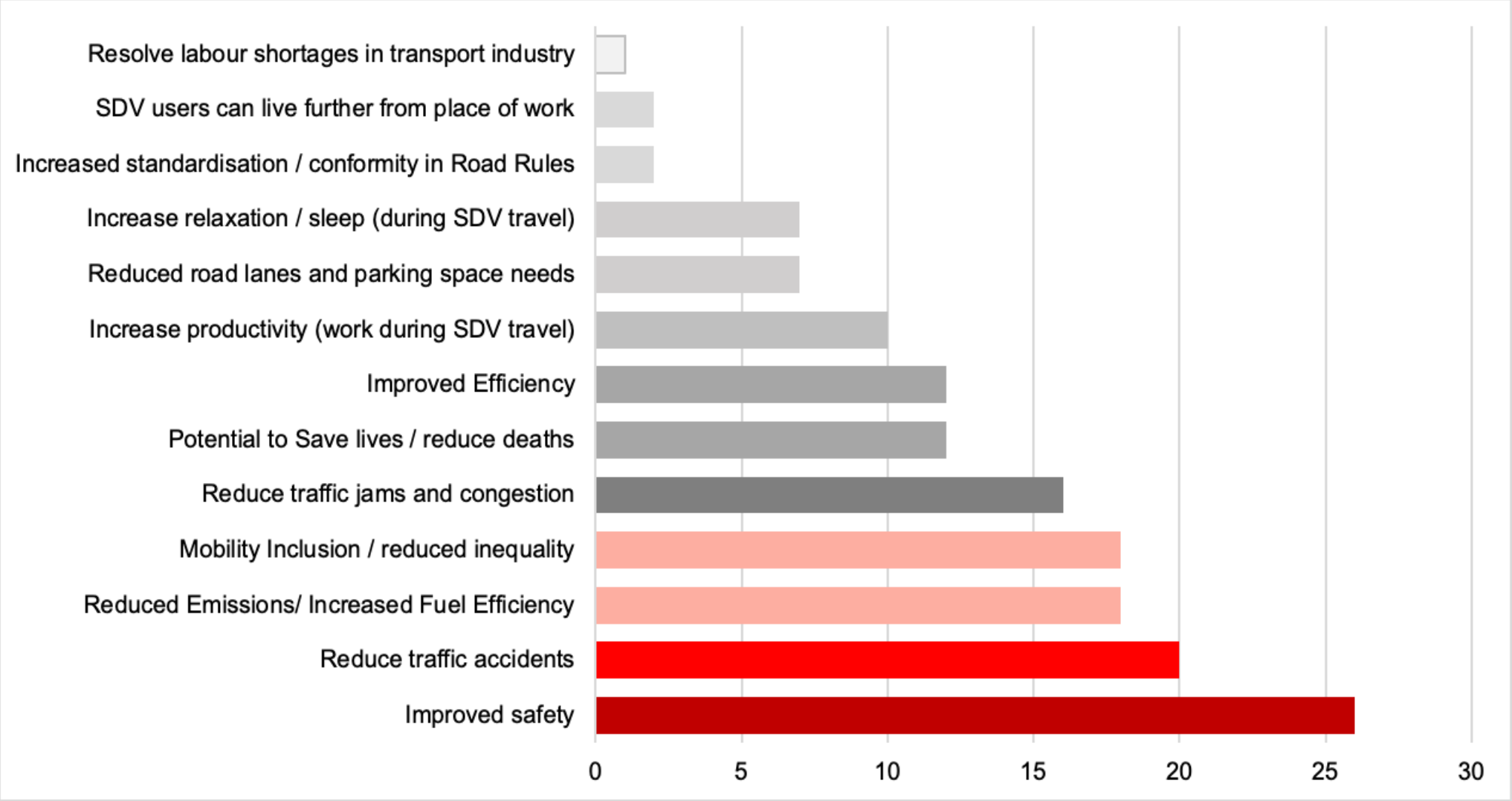

*Figure 6: Proposed benefits from adoption of SDC*

### 4.4.2 Barriers

While six works (15%) proposed no benefits [72, 51, 81, 43, 80, 50], every work in the collection discussed one or more barriers that authors said were inhibiting ubiquitous adoption of SDC. A total of 29 barriers were identified from the literature, as shown in Figure 7. While different causal mechanisms were discussed, 26 works (63%) described the potential for SDC to be involved in or be the source of traffic accidents. Their potential for accidents was a significant motivating factor in the 66% of all works that discussed the potentially difficult task of resolving liability when vehicles without a human driver are involved in collisions. Cybersecurity and the potential for SDC technology to be hacked was identified with the same frequency as the negative impact of law designed for vehicles with a human driver (49%). The latter was not the only barrier arising out of the impact of law, with authors describing issues for development and training of SDC arising out of the *lack of conformity* in traffic laws and road rules across jurisdictions (44%), and the *lack of a legal framework* to support SDC development and implementation in many jurisdictions (17%).

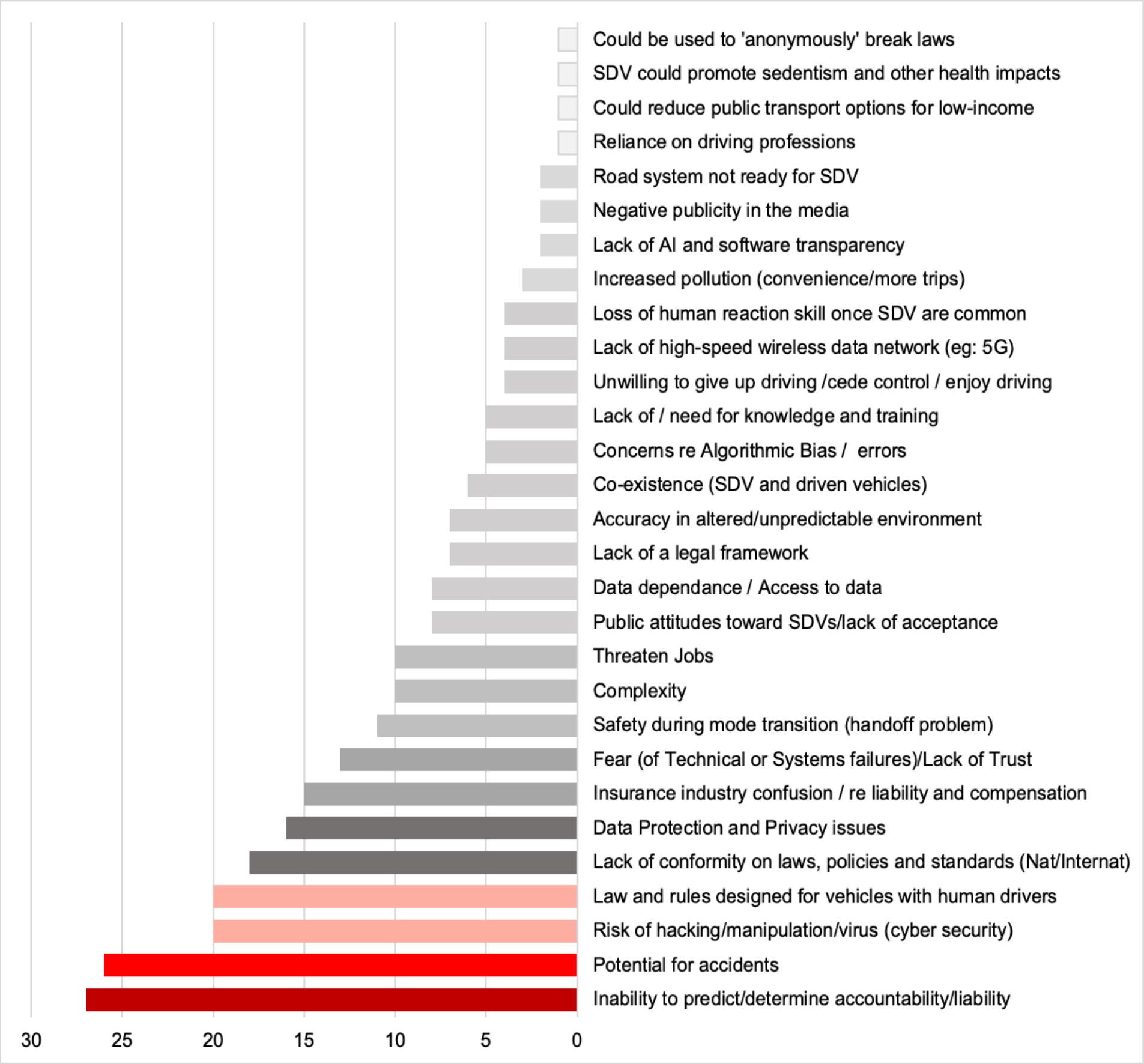

*Figure 7: Barriers preventing adoption of SDC*

### 4.4.3 Facilitating factors

Thirty-five papers (85%) collectively proposed nineteen items listed in Figure 8 for facilitating development or adoption of SDC. Focused directly on the implementation and integration of SDC, four items received more than half of this attention (52%): (i) new laws or policies to enable SDC; (ii) SDC testing on public roads; (iii) new training for regulators, drivers and consumers; and (iv) that SDC manufacturers should accept liability for accidents involving their SDC. One author felt the theoretical risk of SDC misuse in the commission of crimes was significant; proposing integrated biometric identity verification of occupants as the facilitating response in contrast to the much greater focus of other authors on privacy and security concerns. [79]. Those authors focused attention on the data-centric issues SDC bring, which they variously proposed could be resolved through implementation of new infrastructure for data collection (15%) and information sharing (15%), and new privacy and cyber security approaches to protect both the SDC and the data it generates (17%).

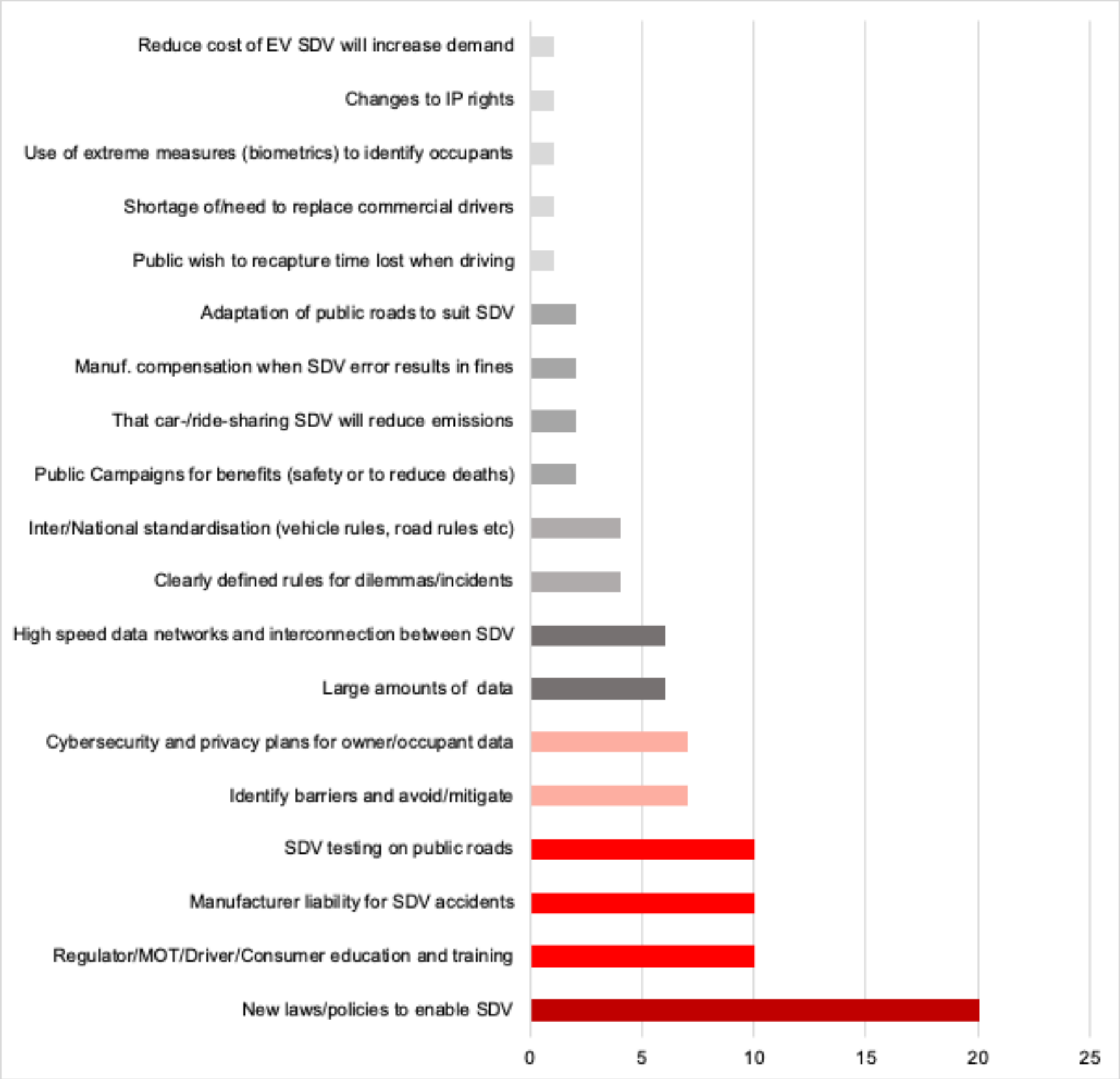

*Figure 8: Facilitating factors for adoption of SDC*

# 4.5 Law and Legislation

## 4.5.1 Existing legislation

Authors identified a wide range of existing legal domains as impacting on SDC, as shown in Figure 9. Much of this legislation naturally regulates SDC in the same way as non-SDC during its design and manufacture (safety standards - 32%), sale (consumer and product law - 46%) and use (registration - 5%, road safety and traffic law - 18%). However, reliance on invention protection (IP, Patent and Trade Secrets law - 15%), questions around the application of existing law when an accident occurs (negligence - 22%, liability - 41%, and insurance law - 29%), and specific laws previously created to scaffold the safety and testing of SDC on populated roads were also discussed. There were also significant discussions regarding the operation of privacy and data protection laws (39%) and eight works (20%) discussed potential for SDC to be involved in crime, including through instruction by users to act as a remote accessory in commission of murder (cite).

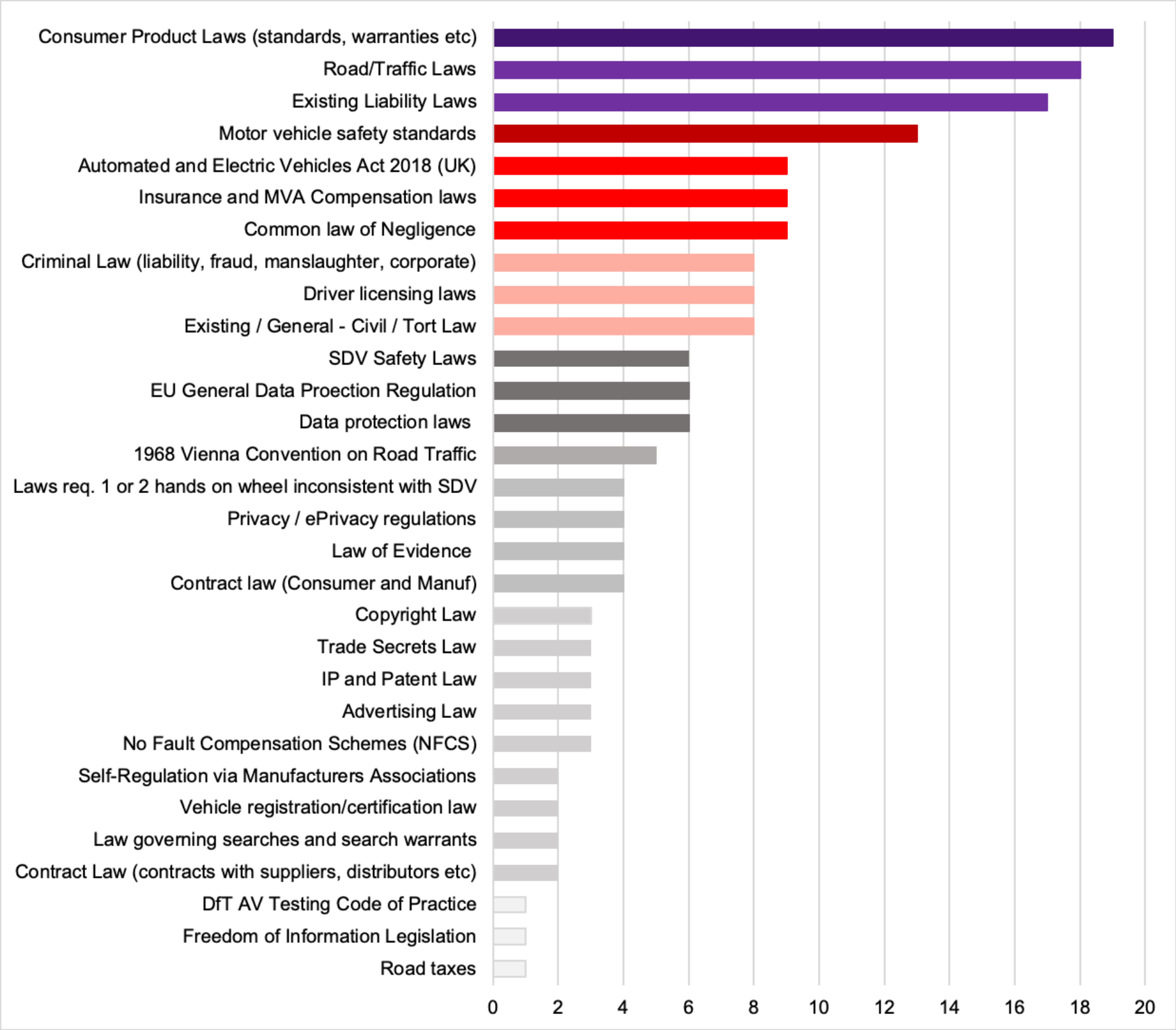

*Figure 9: Existing law regulating SDC*

## 4.5.2 Proposed new legislation

Given *law* and *legislation* were search terms and necessary characteristics for inclusion in this review, it is not surprising that every work provided some proposal either for reform of existing law, or development of entirely new legislation. Shown in Figure 10, motor vehicle accident liability regulation came third in discussions relating to existing law, but were nominated for reform in slightly more than half of all works (54%). Mention of liability was usually accompanied by discourse regarding insurance; potential issues for insurers and claimants when SDC were involved in accidents, how insurers might attribute liability and compensate victims, and whether (cite), or not (cite), legislation to establish manufacturer-funded *no fault compensation schemes* (NFCS) may be one solution capable of resolving these issues. In examining the vast data and connectivity needs of SDC, data protection and privacy concerns were identified by 16 authors; with 13 elevating those concerns to become the second-most identified topic for future legislative attention. These ranged from calls for new legislative approaches to specifically protect occupants from potential exposure of their information to other vehicle users, manufacturers, and third parties like service technicians and police (cite), through to two works that considered amendments were necessary to protect the privacy of pedestrians, bystanders and other road users whose likeness and time-delimited location details would be captured and recorded by SDC sensors and cameras (cite).

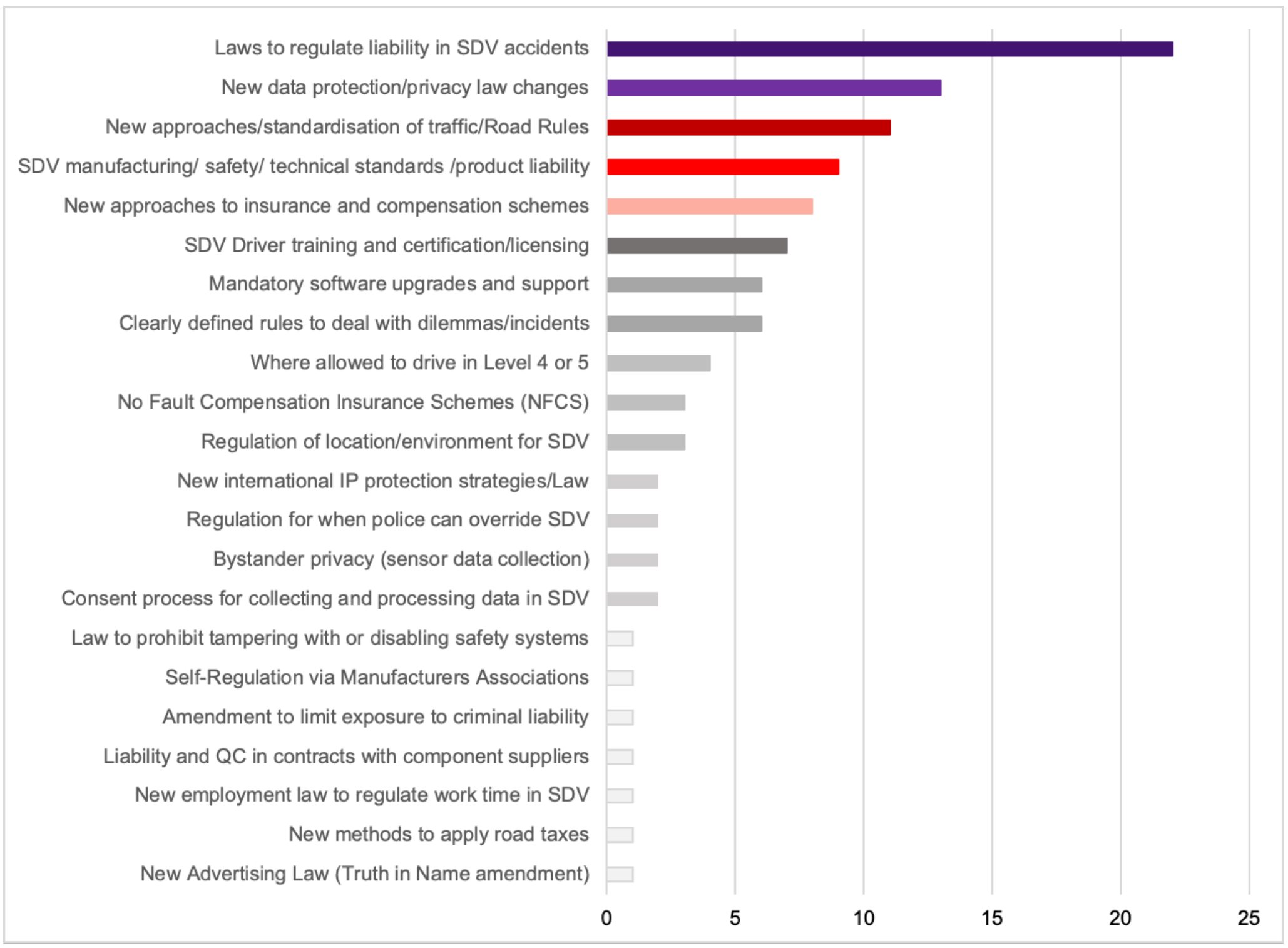

*Figure 10: Future law proposed to enable SDC*

# 5. Discussion

## 5.1 Benefits, barriers and facilitating factors

Using language that strongly supported or reinforced existence of the benefits described in Figure 6, most works promoted benefits that would come from SDC adoption; with some waxing lyrical about positive *radical changes* they posited SDC would make to the way we live [55]. Presentation of presumed benefits generally outweighed discussion of negative social or personal aspects for SDC adoption [49, 18, 40]. While for others, discussion of benefits was a brief distraction during the introduction of their main topic [61, 71, 68]. Authors generally provided citations for their claimed benefits. Some works referenced their benefits to online newspapers like WIRED [46], Business Insider [59], Associated Press and The Verge [79]. Others referenced benefits not to reputable scientific literature that investigated benefits through experiment and empirical analysis, but to government transportation department (NHTSA and UK DoT) and industry-funded economic thinktank reports (McKinsey and RAND) whose writings were profit motivated and focused more on selling the concept of SDC and public road testing to legislators and policy-makers [42, 60, 53, 40, 57, 52]. These reports did not scientifically demonstrate existence of the benefits they endorsed, and carefully couched discussion of them with non-committal words like *could*, *might* and *it is possible that*. Also, potentially resulting from poor and outdated training practices and individual's motivations for profit, recent news articles suggest companies behind SDC development may continue to frustrate their own ability to realise specific benefits[18], such as the often-claimed increased *inclusive mobility* and *reduced transport inequality* for disabled persons we are told these solutions are going to provide [82, 83]. Finally, a small number of works failed entirely to reference the benefits they described [61, 66]. Only four works acknowledged the tenuous and indeterminate nature of SDC benefits. While self-assured of the potential for SDC to reduce traffic deaths, Barrett acknowledged that many of the other benefits raised in her work were less certain [49]. Browne conceded some benefits may be purely optimistic [55], while Rojas-Rueda et al recognised that the potential for benefits was dependent on an indeterminate range of factors [52]. Finally, only one work accurately identified that discussion of benefits was entirely speculative due to the currently limited availability of SDC [63].

[18] Uber and Lyft are two of the companies racing to develop and release SDC. There are many reports of Uber and Lyft drivers refusing to allow disabled passengers to ride in their vehicles and lawsuits have been filed against them on behalf of disabled people in several US states. In one March 2021 San Francisco arbitration decision Uber was ordered to pay Lisa Irving, a blind Californian woman, US$1.1Mill after the Arbitrator found that Uber had been actively educating drivers in approaches they could use to deny disabled persons rides without triggering anti-discrimination laws.

It was claimed that SDC benefits like *reduced traffic accidents* and *deaths*, and thereby *improved safety*, could be substantiated through observance of existing reductions in traffic accident and death statistics in the decade since introduction of Advanced Driver Assist Systems (ADAS) tools like *forward collision mitigation*, *blindspot detection* and *lane change alerts* [57]. However, unmeasured confounding variables like regulation requiring occupants to wear restraints and tougher motor vehicle manufacturing and safety standards, along with annual increases in traffic accidents [84] and road deaths [85, 86], make the conviction with which such assertions were made, at best, questionable.

As is common for any new computing technology where failure could result in loss of life and significant property damage [87], the safety and security of SDC has been identified as an important issue for road users, pedestrians and property owners; whether the vehicle is operating in partial or fully automated mode [79, 70]. Even manufacturers [71] have acknowledged the inevitability of SDC causing the very accidents proponents of SDC have told us they will prevent [56, 42, 73, 74]. While the public have expressed a fear of SDC technical or systems failures (the *fear* barrier) [60, 70], it is also unlikely that vehicle occupants will be aware of such failure even as it is happening [74]. Authors identified many potential causes for SDC failure, including: unpredictable road or weather conditions (the *unpredictable environment* barrier) [73, 81]; mechanical fault (the *reliability* barrier) [73]; manual, partially automated and fully autonomous cars sharing the same roads (the *co-existence* barrier) [42, 73, 79]; the highly technical, complicated and unpredictability of SDC software and systems (the *complexity* barrier) [18, 55, 73, 71]; that SDC computer systems will slavishly obey programmed road rules even when strict adherence may put the vehicle and its occupants in danger (the *inflexibility* barrier) [55]; or issues of vehicle cybersecurity that expose SDC systems to unlawful interference (the *hacking* barrier) [78, 73, 60, 47, 70].

The potential for criminal manipulation, or hacking, was identified in 20 works. One conjecture was that the highly integrated nature of SDC systems would expose significant vulnerabilities that would provide entry points for smart vehicle hacking [47, 65]. However, it has already been established that, using elementary exploits, hackers can take control of non-SDC [88, 89].

Another concern was that it was feasible highly complex software enabling SDC functionality would contain errors and make potentially life-threatening mistakes [73, 67]. While considered to be black swan events[19] [90], automated safety systems taking control at inopportune moments has resulted in many accidents and numerous deaths. Automated fly-by-wire flight telematics systems installed on the original Airbus A320 and more recent A330 aircraft have contributed to numerous accidents where these systems have received errant sensor input and become confused, or misinterpreted the pilot's intentions[20] [91, 92]. In cars, runaway throttle and uncommanded braking issues have also already arisen in the *drive-by-wire* electronic systems of non-SDC vehicles manufactured during the last decade [93].

Privacy concerns have been raised regarding the extensive data collected by sensors and other systems within SDC which may be collected by manufacturers, insurers and the police and used against owners or users [54]. Vehicles like the current Tesla™ range are being promoted as a two-and-a-half tonne PlayStation[21], which hints at the fact that, like gaming consoles, they are gathering and storing data and are capable of using high-speed data networks to share large amounts of potentially sensitive and personally identifying information about owners, passengers, the locations they frequent, and when and how they get there.

We contend that a broader view on SDC technical issues is needed. One where software and hardware assurance, cybersecurity, privacy and data protection are all given equal significance, and where only the most robustly validated solutions are certified for open use.

Aside from authors whose works promoted enabling legislation for SDC and who made up more than half of all facilitating factors, presentation of novel approaches to increase adoption of SDC was generally lacking.

---

[19] Nassim Nicholas Taleb describes Black Swan events as having three things in common. They are: (i) rare; (ii) unexpected; and (iii) their impact is severe.

[20] Flights where investigations have inferred or concluded that the automated control systems played a role in the accident include: Air Asia flight QZ8501 that crashed in the Java Sea off Borneo in 2014, Air France flight 447 that crashed off the coast of Brazil in 2007, the Air New Zealand test flight that crashed into the Mediterranean Sea in 2008, Air France flight 296 that crashed near Mulhouse-Habsheim Airport on the plane's first passenger flight in 1988, and flight LH2904 that crashed at Warsaw Airport in 1993.

[21] In a 2020 episode of The Grand Tour, after having demonstrated the various games and novelty easter-eggs like *celebration mode* that are embedded in the Tesla Model X's integrated software platform, host Jeremy Clarkson describes the Tesla[tm] Autopilot-enabled SDC as "a two-and-a-half tonne, seven-seater PlayStation".

Generally, the primary focus was legislation for identifying or prescribing liability when SDC are involved in or the cause of a traffic accident [18, 55, 78, 56, 42, 60, 75, 71, 79, 54], with suggested mitigations ranging from manufacturers accepting liability when their vehicles are involved in accidents [18, 55, 58], no fault compensation schemes [58, 50], and the incredible suggestion that SDC adoption would increase if manufacturers were insulated from liability to the same degree as vaccine manufacturers; with legislation and something akin to a vaccine court which the author claimed was necessary due to SDC's as-yet unproven *social utility* [18]. The degree of coverage for liability suggests either a major unaddressed issue, or an obsession that has overtaken the entire narrative of the SDC domain to the detriment of other issues of potentially greater significance to users, such as privacy and cybersecurity.

## 5.2 Current law and future legislation

***Current Law:*** Legislation has acted variously to either accelerate or constrain adoption of ADAS technology. For example, even though ABS had been installed on select production vehicles since 1971 [94], from 2004 in Europe and 2013 in the United States of America (USA) legislation made ABS a compulsory safety technology for all motor vehicles being manufactured and sold [95]. Recent legislation in some jurisdictions has also made a small number of other primarily safety-related ADAS compulsory, including rear parking sensors, back-up cameras and forward collision mitigation [96-98]. The EU recently passed regulation that will make fifteen additional safety systems mandatory on European vehicles from 2022, including six of the ADAS technologies listed in Table 1 [99]. However, while UN regulation on some ADAS technologies such as ECE/TRANS/WP.29/2020/81 for automated lane keeping and centring systems was already approved and comes into force in the EU in 2021, use of lane keeping technology is presently restricted in jurisdictions like the UK and was the subject of a recent UK consultation with proposed regulation to allow it at speeds of up to 60km/h remaining on hold [100].

Shown in Figure 11, a wide range of legal domains were directly or implicitly discussed in the context of SDC. *Consumer rights* and *design standards* received considerable attention, mostly focusing on consumer protection in circumstances where SDC are found defective or fail to meet required design or manufacturing standards [56, 76, 71, 59, 70]. Discourse on *road rules* was generally concerned with public, road user and SDC occupant safety [55, 63, 79, 62], while upgrading and standardisation of road rules and traffic signs and how the directives of traffic legislation might be integrated into and complied with by SDC systems were also contemplated [68, 51, 79, 43]. However, most works gave the bulk of attention even on these domains to a single dominant question: apportionment of *liability* when SDC are involved in traffic accidents. The wealth of discussion by authors on this single topic suggesting the need for regulation to resolve liability was overwhelming; even while other authors contemporaneously contended that need for such regulation was receiving very little attention in the literature [55]. Several authors focused on how existing liability legislation operates for non-SDC, in that there is an *automatic presumption* that the human driver is liable until another contributing party is identified [42, 68, 50]. However, numerous authors argued new liability legislation was necessary since fully automated SDC would not have a human driver, with SDC manufacturers often suggested as the appropriate *automatic presumption* replacement [18, 58, 59, 50, 62]. One motivation for this was the need to ensure that compensation could be guaranteed for those injured in accidents caused by SDC without need for protracted litigation of multiple parties, which potentially include the SDC owner, route-commanding occupants, insurer, manufacturer and others in the SDC supply chain. Some authors suggested laws enabling *no-fault compensation schemes* (NFCS) as an insurance approach capable of ensuring immediate compensation for those harmed while the various parties considered or litigated liability [63, 58, 50]. However, other authors were clearly against such approaches, with one unconstructively comparing NFCS with his own particularly negative views on the USA's *Affordable Care Act* (OBAMACARE) [42]. A final work suggested little or no new legislation was required: compared SDC to surgical robots and imputing the latter had seen only limited regulation[22] [56].

[22] The author of that work failed to note that *robotically-assisted surgical devices* (RASD) are overseen by both *medical device* AND *medical device software* regulation. For example, the FDA and TGA classify RASD as Class II (510K or b: medium-high risk) devices requiring extensive assessment and validation prior to authorisation for use with patients, and requiring of ongoing safety assessment once the device is clinically available. Further, their use is also impacted by the need for practitioner *scope of practice* certification, data protection and privacy regulation, liability and tort law, malpractice and insurance law, software assurance standards and a range of other laws, regulations and codes of practice common to both safety-critical technology and medical devices.

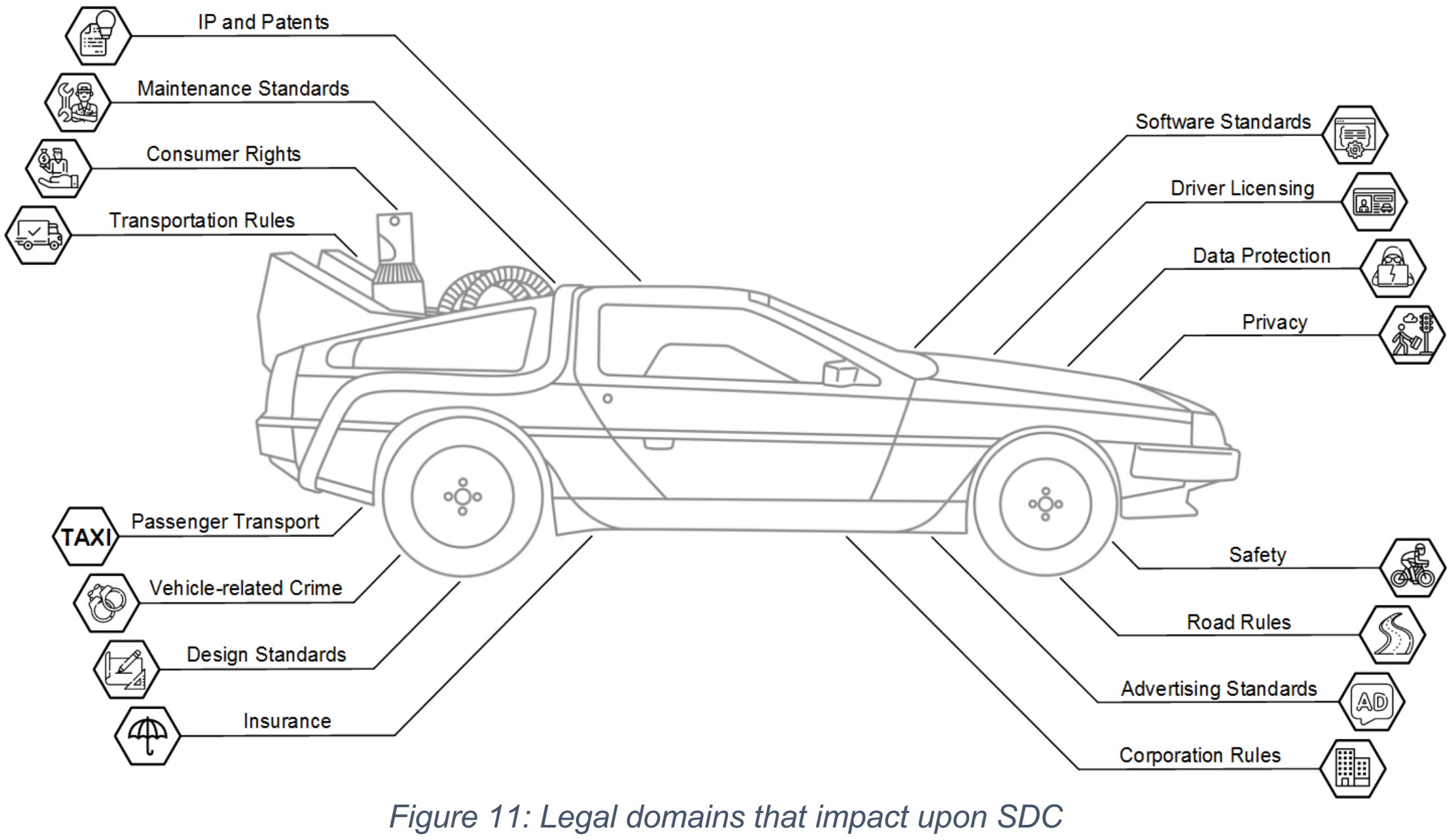

*Figure 11: Legal domains that impact upon SDC*

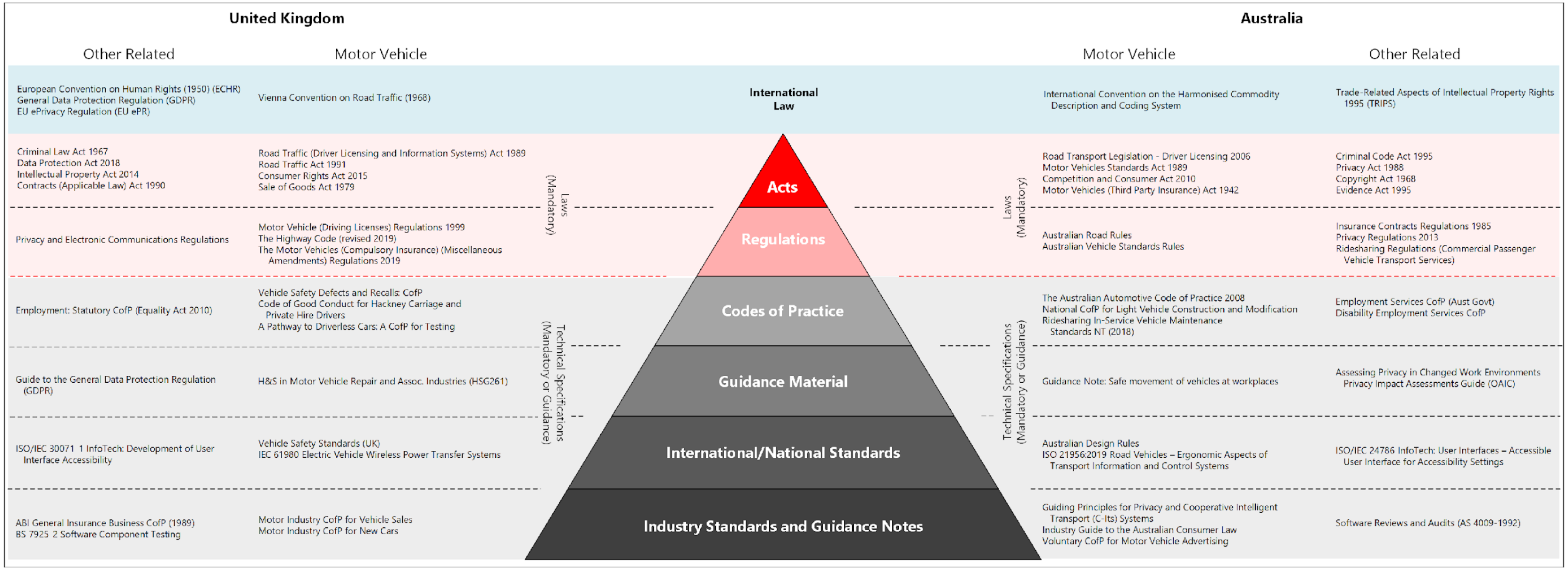

*Figure 12: Levels of Regulation for SDC*

It is common for related laws, regulations, policies and standards that collectively govern our modern world to be visualised using a *pyramid* structure. Rising from the bottom, each layer represents the degree of significance each element has in the governance structure, with Acts of Parliament the most significant being represented at the peak. Also, the increasing width of each layer as one works down from the peak can also represent the volume of content that exists at that layer. Such a pyramid is illustrated at the centre of Figure 12, in which it is used as a framework summarising a range of laws and regulations governing motor vehicles, including self-driving cars, across two jurisdictions, namely, the United Kingdom and Australia. The laws, regulations and standards that directly relate to motor vehicles are presented nearer to the pyramid and at the appropriate layer for the type of laws. It is important to note that the degree to which the laws, regulations and standards are mandatory should be seen to be decreasing as one goes down the pyramid's base where they also tend to become technical specifications aimed at achieving the effect of the mandatory laws and regulations at the top of the pyramid. The laws, regulations and standards identified at each layer are the most significant. Thus, Figure 12 is not an exhaustive summary of the laws, regulations and standards for motor vehicles in both the UK and Australia. Other related laws, regulations and standards are provided to illustrate the fact that there are many other laws related to motor vehicles other than those that are directly targeted at motor vehicles. What is clear for both of the two jurisdictions is the absence of laws and regulations that are specific to the self-driving car. There may be the assumption that existing laws, regulations and standards are generic enough to be applicable, extendable, adaptable, interpretable or extrapolatable to the case of self-driving cars. However, this could only be partial and incomplete as self-driving cars are a new technology with strong possibility of *emergent* operational and legal properties, especially arising from embedded AI with decision-making capability that alters the behaviour of the car on the road.

# 6. Conclusion

From international regulations that serve to expedite homologation of vehicles through to country-specific road rules, motor vehicles are extensively regulated. This work proposed a taxonomy for understanding existing *automated driving assistant system* (ADAS) features prior to reviewing a collection of literature providing law-focused analysis and discussion of self-driving cars (SDC). We found that many recent and proposed laws and regulations are predominately based on thirteen claimed benefits, yet those benefits are purely speculative and untested. While barriers permeate within almost every area of technology, society, laws and regulation, the primary focus of many authors were the inter-related issues of negligence, liability and blame apportionment. Also, little thought has been given to identifying the real risks that SDC may bring, and the likelihood and impact of their occurrence. And finally, application of new and potentially more complicated law and regulation was most frequently described as the primary approach for resolving issues and enabling broad adoption of SDC.

Legal scholarship regarding AI generally and SDC specifically must look more broadly than negligence and liability if we are to prepare the general public for coexistence, and eventual full adoption of SDC. Results of this paper are significant in that they point to the need for deeper comprehension of the broad impact of all existing law and regulations on the introduction of SDC technology, with a focus on identifying only those areas truly requiring ongoing legislative attention.

**Acknowledgements**

The authors acknowledge support for this research from the following sources: From EPSRC under project *AISEC: AI Secure and explainable by Construction* - EP/T026952/1; and RAEng under project *SafeAIR: Safer aviation from ethical Autonomous Intelligence Regulation* - ICRF2122-5-234.